\documentclass[preprint,12pt,authoryear]{elsarticle}

\usepackage{lineno,hyperref}
\modulolinenumbers[5]
\usepackage{algorithmic} 
\usepackage{algorithm}
\usepackage{graphicx}
\usepackage{multirow}    
\usepackage{hyperref}
\usepackage{geometry}
\usepackage{float}
\usepackage{subfloat} 

\usepackage{natbib}
\usepackage{amsmath}
\usepackage{url}
\usepackage{lineno,hyperref}
\modulolinenumbers[5]
\usepackage{color}
\usepackage{graphicx} 
\usepackage{subfig} 

\date{}
\usepackage{amssymb}
\usepackage{natbib}
\setcitestyle{numbers,square}
\usepackage{threeparttable}

\usepackage{subfloat}
\usepackage{stfloats}

\usepackage{algorithmic}
\usepackage{algorithm}
\usepackage[algo2e,ruled,vlined]{algorithm2e}

\begin{document}

\begin{frontmatter}

\title{$L_p$-norm Distortion-Efficient Adversarial Attack}


\author[1]{Chao Zhou\corref{cor1}}
\cortext[cor1]{This work was submitted to Signal Processing: Image Communications in December 2022.}

\author[1]{Yuan-Gen Wang\corref{cor2}}
\cortext[cor2]{Corresponding author: }
\ead{wangyg@gzhu.edu.cn}
\author[1]{Zi-Jia Wang}
\author[]{Xiangui Kang$^{2}$}

\address{$^{1}$School of Computer Science and Cyber Engineering, Guangzhou University, China\\
$^{2}$School of Computer Science and Engineering, Sun Yat-Sen University, China}

\begin{abstract}
Adversarial examples have shown a powerful ability to make a well-trained model misclassified. Current mainstream adversarial attack methods only consider one of the distortions among $L_0$-norm, $L_2$-norm, and $L_\infty$-norm. $L_0$-norm based methods cause large modification on a single pixel, resulting in naked-eye visible detection, while $L_2$-norm and $L_\infty$-norm based methods suffer from weak robustness against adversarial defense since they always diffuse tiny perturbations to all pixels. A more realistic adversarial perturbation should be sparse and imperceptible. In this paper, we propose a novel $L_p$-norm distortion-efficient adversarial attack, which not only owns the least $L_2$-norm (or $L_\infty$-norm) loss but also significantly reduces the $L_0$-norm distortion. To this aim, we design a new optimization scheme, which first optimizes an initial adversarial perturbation under $L_2$-norm (or $L_\infty$-norm) constraint, and then constructs a dimension unimportance matrix for the initial perturbation. Such a dimension unimportance matrix can indicate the adversarial unimportance of each dimension of the initial perturbation. Furthermore, we introduce a new concept of adversarial threshold for the dimension unimportance matrix. The dimensions of the initial perturbation whose unimportance is higher than the threshold will be all set to zero, greatly decreasing the $L_0$-norm distortion. Experimental results on three benchmark datasets (MNIST, CIFAR10, and ImageNet) show that under the same query budget, the adversarial examples generated by our method have lower $L_0$-norm and $L_2$-norm (or $L_\infty$-norm) distortion than the state-of-the-art. Especially for the MNIST dataset, our attack reduces 8.1$\%$ $L_2$-norm distortion meanwhile remaining 47$\%$ pixels unattacked. This demonstrates the superiority of the proposed method over its competitors in terms of adversarial robustness and visual imperceptibility.
\end{abstract}


\end{frontmatter}


\section{Introduction}
Nowadays, deep convolutional neural networks (CNNs) have achieved great success in computer vision tasks, such as image classification, image enhancement, and image quality assessment. However, recent studies \cite{HE2022116747}, \cite{szegedy2014intriguing}, \cite{LIANG2022116659} have shown that the absurdly wrong decisions will be made by the well-trained CNN classifiers when the images are added with imperceptible disturbances. These images that mislead classifiers are called adversarial examples \cite{szegedy2014intriguing}. Thus, the application of CNN models is restricted due to the existence of adversarial examples, especially in some security-sensitive fields. To improve the robustness of the CNN models and reduce the risk of adversarial examples, many researchers have devoted a good deal of attention to the field of adversarial attacks.
According to whether the inner parameters of the target model are access to users, adversarial attacks can be classified into white-box attacks and black-box attacks. The white-box attacks are performed when all parameters of the target model are exposed to users. By contrast, the black-box attackers can only obtain the model prediction results through the application interface. Especially for a hard-label system, the attackers can only know the one-hot output vector. In practice, for security and business reasons, the model's internal parameters are often not public to users. This paper focuses on the hard-label black-box attacks, which are more practical for real-world CNN models. 

\begin{figure*}[tp]
	\centering	\includegraphics[width=1\textwidth]{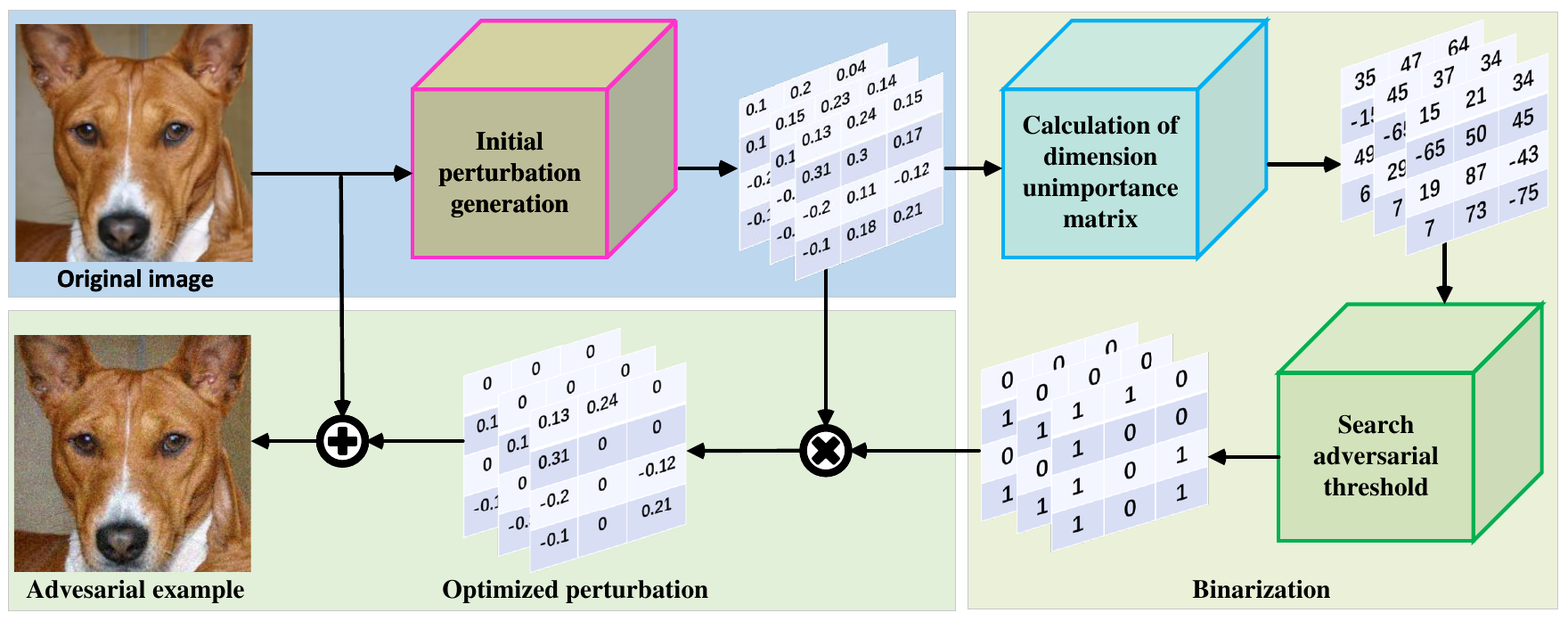}
	\caption{Overview of the proposed method, which includes the initial perturbation generation, the calculation of dimension unimportance matrix, and the perturbation dimension optimization.}
	\vspace{-0.5cm}
\end{figure*}

In current black-box attacks, the most commonly used metrics for adversarial examples are $L_2$-norm, $L_0$-norm, and $L_\infty$-norm. $L_2$-norm attacks mainly include gradient estimation methods and transferability attack methods. This type of attack method estimates the gradient of an image in various ways, and then adds adversarial perturbations to the direction of the estimated gradient. Although $L_2$-norm attacks can optimize the total Euclidean distortion of the adversarial example, they don't care about the distribution of the perturbation and always lead to a global perturbation. There are two drawbacks of global perturbation. One is that there are too many useless pixel modifications \cite{xiang2021local}, \cite{zhou2022object}. The other is that it suffers from weak robustness against adversarial defenses such as general image processing-based defense methods since they often diffuse tiny perturbations to all pixels \cite{dong2020robust}. Therefore, there are some limitations in taking $L_2$-norm loss as the optimization goal only. Unlike $L_2$-norm attacks, $L_0$-norm attacks optimize the number of pixels that must be changed when generating an adversarial example. The less the number of changed pixels, the better the adversarial example. However, the $L_0$-norm attacks have no restriction on modification of each pixel, causing the perturbed pixels to be easily visible. $L_\infty$-norm attacks limit the maximum change of a single pixel. However, such attacks must change every pixel and color channel, which seems unrealistic and hard to realize.

As reviewed, the recurrent adversarial examples always have their own defects using only one metric. In this paper, we propose a new optimization scheme that considers both $L_2$-norm (or $L_\infty$-norm) loss and $L_0$-norm loss simultaneously. To this end, we first generate an initial perturbation which is optimized on $L_2$-norm (or $L_\infty$-norm). Then, we design a new algorithm to calculate a dimension unimportance matrix for the initial perturbation, which can indicate the adversarial unimportance of each dimension of the initial perturbation. Moreover, we introduce a new concept of adversarial threshold and use a binary search algorithm to search the threshold value. On this basis, we set the dimensions in the initial perturbation whose unimportance values exceed the threshold to zero. As a result, we can get adversarial examples that are optimized on $L_p$-norm metrics. Extensive experimental results demonstrate the effectiveness of the proposed method. We believe this work poses a new challenge to the robustness of the CNN classifiers, shedding some light on the defense against adversarial attacks. The major contributions of this paper can be summarized as follows:

\begin{itemize}
	\item We propose a new optimization scheme to generate the adversarial example under $L_p$-norm restriction, resulting in a minimal $L_0$-norm distortion meanwhile the $L_2$-norm (or $L_\infty$-norm) distortion is also minimum.
		
	\item We construct a dimension unimportance matrix that can indicate the adversarial unimportance of each dimension in the initial perturbation, helping benefit the optimization process.
	
	\item We introduce a new concept of adversarial threshold. Based on this, we can set the dimensions of the initial perturbation whose unimportance value are higher than the threshold to zero, significantly reducing the $L_0$-norm distortion.
	
	\item Extensive experiments show that under the same query budget, the adversarial examples generated by our method achieve both lower $L_2$-norm and $L_0$-norm distortion than the state-of-the-art methods, especially the obvious advantage on the MNIST dataset.
\end{itemize}

The rest of this paper is organized as follows. Section 2 summarizes the existing mainstream methods of generating adversarial examples. Section 3 introduces our method in detail and then provides the pseudocode implementation of each algorithm part. Experimental results and analysis are shown in Section 4. We make a conclusion in Section 5 by summarizing our work and discussing future works.

\section{Related Works}
The task of image classification is to successfully predict what humans can see in images. The study of adversarial examples involves the robustness of CNN models by inputting small changes. A hostile attacker usually adds perturbations in the images to fool the CNN classifier. Since the changes are often carefully designed by attackers, they are very small and even imperceptible to humans. The robustness problem of neural networks was first discovered by Szegedy \textit{et al.} \cite{szegedy2014intriguing} in 2013. Accordingly, the concept of adversarial example was proposed, usually applied in attacking image classification models. In general, adversarial attacks can be divided into targeted attacks and untargeted attacks. The targeted attacks require the classifier to predict the image as a specified error class, while the untargeted attacks only need to fool the model and make a wrong decision about the input image. Based on the information accessed by attackers, adversarial attacks are mainly divided into white-box attacks and black-box attacks.

\subsection{White-box Attacks}
In the beginning, many works have been proposed for white-box attacks, where the attackers need to access the full parameters of CNN classifiers, including network structure and weights. In \cite{GoodfellowSS14}, Goodflow \textit{et al.} proposed a fast gradient sign method (FGSM). Kurakin \textit{et al.} \cite{kurakin2016adversarial}, Chen \textit{et al.} \cite{chen2018ead}, and Madry \textit{et al.} \cite{madry2017towards} also proposed related algorithms based on gradient computation. The C\&W attack proposed by Carlini and Wagner \cite{carlini2017towards} is the first to deal with the adversarial attack problem from an optimization perspective. Athalye \textit{et al.} \cite{athalye2018obfuscated} proposed the BPDA attack, which fooled some models designed with gradient obfuscated processing and successfully bypassed many defense methods. Unlike the typical attacks based on the minimization $L_p$-norm perturbations, Zhang \textit{et al.} \cite{zhang2019limitations} exploited scaling and shifting to generate perturbations. In \cite{MARRA2018240}, Marra \textit{et al.} studied the influence of adversarial examples on CNN-based methods for camera model identification. The success rate of white-box attacks is usually high. However, it is not practical because it depends heavily on the knowledge of the internal parameters of the target model, which is often difficult to obtain in reality. 

\subsection{Black-box Attacks}
In contrast to white-box attacks, black-box attacks can only query the model's output. The existing black-box attacks can be divided into transfer attacks and query-based attacks. Since the internal information of the CNN model is unknown, some attackers first choose to train another model, called the substitute model, and then perform the white-box attack through the substitute model whose internal parameters are accessed. Papernot \textit{et al.} \cite{papernot2017practical} proposed to train a substitute model by querying the output of model for the first time. Although some works have appeared in the field of transfer attack, we focus on the query-based attacks in the paper. According to the output forms, the CNN models can be divided to soft-label and hard-label settings. For the soft-label setting, the model outputs both the classification decision and the corresponding probability. Chen \textit{et al.} \cite{chen2017zoo} proposed a zeroth order optimization (ZOO) framework that uses a finite difference method to approximately estimate the gradient. Ilyas \textit{et al.} \cite{ilyas2018black} proposed to use neural evolution strategy (NES) directly. Inspired by this, many variants were proposed to promote the development of adversarial attacks further. For example, Tu \textit{et al.} \cite{tu2019autozoom} proposed the AutoZOOM framework using an adaptive random gradient estimation strategy and dimension reduction technique. Ilyas \textit{et al.} \cite{ilyas2019prior} exploited prior information of the gradient with the proposed bandit optimization. Alzantot \textit{et al.} \cite{alzantot2019genattack} proposed a black-box optimizer for the soft-label setting. To reduce the number of queries,  Guo \textit{et al.} \cite{guo2019simple} proposed a simple black-box attack (SimBA), which is implemented by adding a fixed size perturbation to the pixels one by one until the attack succeeds. Ru \textit{et al.} \cite{ru2019bayesopt}  proposed to generate adversarial examples by using the Bayesian optimization in combination with Bayesian model selection and improved the query effectiveness.

In hard-label settings, the model only outputs the classification decision, which is also known as the top-1 prediction class. Instead of the predicted probability, the attackers can only obtain the corresponding decision output (i.e. hard-label) by querying the model. Therefore, the hard-label setting provides less information than the soft-label one. Brendel \textit{et al.} \cite{brendel2018decision} was the first to study a hard-label attack problem and proposed a search algorithm (boundary attack) that randomly walks near the decision boundary to find the minimum perturbation necessary to fool the target model. The boundary attack performs a rejection sampling near the decision boundary. Ilyas \textit{et al.} \cite{ilyas2018black} designed a query-limited attack method to turn the hard label into a soft-label problem by estimating the predicted probability. On the contrary, Cheng \textit{et al.} \cite{cheng2019query} reformulated the hard-label attack as an optimization problem and proposed an Opt-attack method, which is to find a direction that can produce the shortest distance to the decision boundary. Based on the Opt-attack, Cheng \textit{et al.} \cite{ChengSCC0H20} designed a query efficient optimization algorithm named Sign-OPT. It is a new zeroth order optimization algorithm where sign SGD converges fast. Shukla \textit{et al.} \cite{shukla2021simple} proposed an adversarial attack based on a Bayesian Optimization (BO) to search adversarial examples in a structured low dimensional subspace.

\section{Proposed Method}
Denote $f_K(\cdot): R^{M}\to{1,...,K}$ as the black-box multi-class classifier and $x_0$ as the original image respectively, where $M$ is the dimension of the input and $K$ is the number of classes. For a test example $x\in R^{M}$, we can get the predicted category $y$ by 
\begin{equation}
	y =  \underset{k\,=1,...,K}{\textrm{arg max}}\, f_K(x).
\end{equation}
Denote $y_0$ as the ground truth of the original image $x_0$. If $y_0 = \underset{k\,=1,...,K}{\textrm{arg max}}\, f_K(x_0)$, it can be said that the image $x_0$ is classified correctly. For untargeted attack setting, the goal of the adversarial attack is to find an adversarial example $x^*_0 = x_0 + \delta $ subject to the following constraint
\begin{equation}
	\begin{split}
	&\underset{x^*_0\in R^{M}}{\textrm{arg min}}\, d(x^*_0 , x_0),\,\, \\
	& \textrm{s.t.} \,\, \underset{k\,=1,...,K}{\textrm{arg max}}\, f_K(x^*_0) \neq \underset{k\,=1,...,K}{\textrm{arg max}}\, f_K(x_0),
	\end{split}	
\end{equation}
where $d(\cdot, \cdot)$ denotes the distance metric. 

Usually, the most commonly used metrics for adversarial examples are $L_2$-norm, $L_0$-norm, and $L_\infty$-norm. The smaller the distance, the better the adversarial example. $L_2$-norm refers to the Euclidean distance, that is  
\begin{equation}
d(x^*_0, x_0)=\sqrt{\sum_{m=1}^{M}(x^*_{0m} - x_{0m})^2},
\end{equation}
where $x_{0m}$ is the $m^{th}$ pixel of an image. Although $L_2$-norm can calculate the overall distortion of the adversarial example, it does not constrain how the perturbation is distributed and how many pixels are changed. Therefore, taking it as the optimization target will often lead to a global perturbation distributed in the whole image. Different from the $L_2$-norm attack, the target of $L_0$-norm attacks is to minimize the number of attacked pixels when generating an adversarial example.  The fewer changed pixels, the better the adversarial example. The $L_0$-norm distance can be calculated by
\begin{equation}
		d(x^*_0, x_0) = \sum_{m=1}^{M}\textrm{1}_{|x^*_{0m} - x_{0m}|}.
\end{equation}
However, the $L_0$-norm attacks cause the perturbed pixels to be easily visible. The $L_\infty$-norm attacks cause a relatively slight change on each pixel. However, it has to change every pixel and color channel, which seems unrealistic and hard to realize.  $L_\infty$-norm is expressed as
\begin{equation}
	d(x^*_0, x_0) = \underset{m \,=1,...,M}{\textrm{max}}|x^*_{0m} - x_{0m}|.
\end{equation}

\subsection{Attack Framework}
Due to the similar effect of  $L_2$-norm and $L_\infty$-norm, we mainly describe our method with $L_2$-norm in the following.
Since the $L_2$-norm attacks will lead to a global perturbation and the $L_0$-norm attacks are easily naked-eye visible, using a single distance metric cannot achieve a good attack effect. Motivated by this, we propose an optimization framework to find the adversarial example that is optimized on both $L_2$-norm and $L_0$-norm. Our optimization objective function is defined as 
\begin{equation}
	\begin{split}
		&\underset{x^*_0\in R^M}{\textrm{arg min}} 
		\sum_{m=1}^{M}\textrm{1}_{|x^*_{0m} - x_{0m}|},\,\,\\
		& \textrm{s.t.} \,\, \underset{k\,=1,...,K}{\textrm{arg max}}\, f_K(x^*_0) \neq \underset{k\,=1,...,K}{\textrm{arg max}}\, f_K(x_0),\\
		& \textrm{s.t.}\,\, \underset{x^*_0\in R^{M}}{\textrm{arg min}}  
		\sqrt{\sum_{m=1}^{M}(x^*_{0m} - x_{0m})^2}.\,\, 
	\end{split}	
\end{equation}
Different from previous attack methods, our method is to find the adversarial example with the least $L_2$-norm loss and the minimum number of changed pixels. 
Our attack framework consists of three parts: the initial perturbation generation, the dimension unimportance matrix calculation, and the perturbation dimension optimization. The overall framework of our attack method is shown in Figure 1.

\subsection{Generate Initial Perturbation}
We adopt the Sign-Opt attack \cite{ChengSCC0H20} as the decision-based attack baseline to find an initial perturbation. The process consists of two steps. The first step is to find a direction vector whose distance to the decision boundary is the minimum among a set of randomly generated direction vectors that follow the standard normal distribution. The objective function
  $g(\cdot):R^{M}
$ can be expressed as
\begin{equation}
	\underset{\theta}{\textrm{min}}\,\, g(\theta), \,\,  \textrm{where} \,\, g(\theta) = \textrm{arg}\underset{\lambda_0>0}{\textrm{min}}(f_K(x_0+\lambda_0\frac{\theta}{\parallel\theta\parallel}) \neq y_0),
\end{equation}
where $\theta$ is a direction vector, and the $\frac{\theta}{\parallel\theta\parallel}$ is the unit vector of direction $\theta$. $\lambda_0$ denotes the distance from the direction $\frac{\theta}{\parallel\theta\parallel}$ to the decision boundary, and it is evaluated by a binary search algorithm.
By generating a series of $\theta$ and calculating the corresponding $g(\theta)$, we will find an optimized initial direction vector $\theta$ and its corresponding $\lambda_0$.

The second step is to find a direction $\hat{g}$ for updating $\theta$, which satisfies the condition $g(\theta-\eta \hat{g}) < g(\theta)$ in the new direction $\theta-\eta \hat{g}$, where $\eta$ is an automatically adjusted parameter. The direction $\hat{g}$ can be estimated by the sign gradient  method, which can be expressed by

\begin{equation}
	\hat{g}\, = \, \sum_{q=1}^{Q}\textrm{sign}(g(\theta+\epsilon\mu_q)-g(\theta))\mu_q,
\end{equation}
 where $\mu_q$ denotes the $q^{th}$ noise which follows the standard normal Gaussian distribution, and $Q$ is the number of $\mu_q$.

\begin{figure*}[ht]
	\centering	\includegraphics[width=0.5\textwidth]{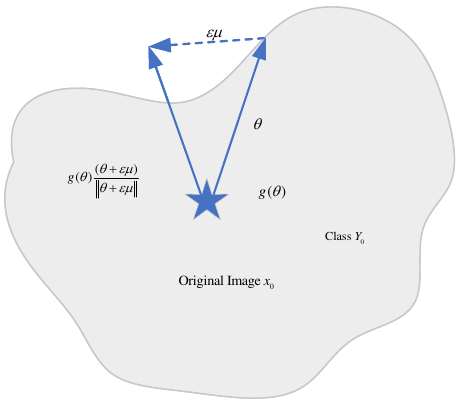}
	\caption{Illustration of generating initial perturbation.}
	\vspace{-0.5cm}
\end{figure*}

\begin{equation}
	\textrm{sign}(g(\theta+\epsilon\mu_q)-g(\theta))=\left\{
	\begin{array}{cl}
		+1,  &  \textrm{if} \, (f_K(x_0+ g(\theta)\frac{(\theta+\epsilon\mu)}{\parallel\theta+\epsilon\mu\parallel})=y_0), \\
		-1,  &  \textrm{otherwise}, \\
	\end{array} \right.
\end{equation}
where $\epsilon$ is a very small smoothing parameter. As shown in Figure 2, for a new direction $\theta+\epsilon\mu$, we select a point at the same distance from the original image $x_0$ in this direction. Then, we test whether the point lies inside or outside the decision boundary. The produced perturbation creates an adversarial example when added to the original image $x_0$. If the produced perturbation is an adversarial example, i.e. $f_K(x_0+ g(\theta)\frac{(\theta+\epsilon\mu)}{\parallel\theta+\epsilon\mu\parallel})\,\,!=y_0$, which means the distance to the decision boundary is smaller for the new direction $\theta+\epsilon\mu$, a smaller value of $g$ can be obtained. Thus, we can see that $\mu$ is a descent direction to minimize $g$.
We iteratively obtain multiple $\hat{g}$, and then calculate each distance of the new direction $\theta-\eta \hat{g}$, i.e. $g(\theta-\eta \hat{g})$. If $g(\theta-\eta \hat{g})< g(\theta)$, we use $\theta-\eta \hat{g}$ to replace $\theta$, and then continue to iterate. In this way, we get the initial perturbation direction $\theta_0$ (the best $\theta$) and the corresponding $\lambda_0$ which are optimized on the $L_2$-norm metric. 

\begin{algorithm2e*}[tp]
	\caption{Calculate the dimension unimportance matrix}
	\label{alg1}
	\begin{algorithmic}[1]
		\REQUIRE Image $x_0$ and ground truth $y_0$, hard-label model $f_K(\cdot)$, initial perturbation direction $\theta_0$ and corresponding $\lambda_0$
		\ENSURE Dimension unimportance matrix $\beta$
		\STATE Initialize list $Lp$, $Ln$, $S_p$, $S_n$ empty
		\FOR{0.00,0.00,1.00{$i=1,2,...,I$ }} 
		\STATE Randomly generate 0/1 matrix $\omega_i $
		\STATE $\theta_0^*$ $\leftarrow$ $\theta_0\cdot\omega_i$ 
		\STATE $\gamma_i \leftarrow R(\omega_i) \cdot\theta^*_0 $
		\IF{$f_K(x_0 + \lambda_0\theta^*_0) \,\,!=y_0$}
		\STATE Symbol matrix $S_i\leftarrow R(\omega_i)$
		\STATE $S_p.\textrm{append}(S_i)$		
		\STATE $\gamma_p.\textrm{append}(L_2(\gamma_i))$	
		\ELSE		
		\STATE Symbol matrix $S_i\leftarrow -R(\omega_i)$
		\STATE $S_n.\textrm{append}(S_i)$			
		\STATE $\gamma_n.\textrm{append}(-L_2(\gamma_i))$			
		\ENDIF
		\ENDFOR 			
		\STATE $\gamma_p$ $\leftarrow$ normalize($\gamma_p$)
		\STATE $\gamma_n$ $\leftarrow$  normalize($\gamma_n$)
		
		\STATE Compute dimension unimportance matrix 
		$\beta \leftarrow$ $\sum_{1}^{\textrm{len}(\gamma_p)}\gamma_p[l]\cdot S_p[l] + \sum_{1}^{\textrm{len}(\gamma_n)}\gamma_n[h]\cdot S_n[h]$						
	\end{algorithmic}
	
\end{algorithm2e*}

\subsection{Calculate Dimension Unimportance Matrix}
Since only little information about the model can be obtained from the hard-label black-box setting, the process of generating the initial perturbation converts to the problem of finding the direction that is the shortest away from the decision boundary. However, it is difficult to estimate $\theta_0$ and $\lambda_0$ accurately. In fact, it is unnecessary to change all pixels of the whole image to generate an adversarial example. 
Inspired by this, we propose a perturbation dimension optimization method by calculating a dimension unimportance matrix that represents the unimportance of each dimension of the optimized initial perturbation. The calculation of the dimension unimportance matrix is shown as follows.

First, we randomly set a part of dimensions of $\theta_0$ to zero in our iteration and obtain $\theta_0^*$ by

\begin{equation}
	\theta^*_0 = \theta_0\cdot\omega_i,
\end{equation}
where $\omega_i$ is a random 0-1 matrix and $i$ means the $i^{th}$ iteration. According to the judgment whether the produced perturbation is still an adversarial example, i.e. $f(x_0+\lambda_0\theta_0^*)\,\,\underset{=}{?}\,\,y_0$, we can construct a symbol matrix $S_i$ as

\begin{equation}
	S_i=\left\{
	\begin{array}{cl}
		R(\omega_i),  &  \textrm{if } \, f(x_0 + \lambda_0\theta^*_0) \,\,!=y_0, \\
		-R(\omega_i),  &  \textrm{otherwise} ,\\
	\end{array} \right.
\end{equation}
where $R(\cdot)$ denotes the operation of flipping the elements in the 0-1 matrix, i.e. changing  0 to 1 and 1 to 0. The value of 1 in $S_i$ indicates that the corresponding dimension in the initial perturbation is unimportant. Then, the dimensions of the initial perturbation can be optimized by setting these dimensions to zero. Conversely, if the value is -1, it means that the dimensions cannot be set to zero. When the produced perturbation is still an adversarial example, we use two lists to store the dimensions data that are set to zero in each iteration. We use a list $\gamma_p$ to record the set of positive examples which can still satisfy $f_K(x_0 + \lambda_0\theta^*_0) \,\,!=y_0$ when set to zero, and a list $S_p$ to record the corresponding positive $S_i$. $\gamma_n$ and $S_n$ are used for the negative examples. $\gamma_i$ is the data of the dimensions in $\theta_0$, which are set to zero in a single iteration. It can be formulated as
\begin{equation}
	\gamma_i = R(\omega_i)\cdot\theta^*_0.
\end{equation}

Second, we calculate the corresponding weight $\alpha$ of the symbol matrix $S_i$. To reduce the $L_2$-norm loss while optimizing the perturbation dimension, we multiply the symbol matrix by a corresponding weight. The weight is calculated by the maximum value and minimum value normalization method, which is expressed as

\begin{equation}
	\begin{split}
		&\alpha_l = \frac{L_2(\gamma_p[l])-\textrm{min}(L_2(\gamma_p))}{\textrm{max}(L_2(\gamma_p))-\textrm{min}(L_2(\gamma_p))},\,\, \\
		& \alpha_h = \frac{(-L_2(\gamma_n[h]))-\textrm{min}(-(L_2(\gamma_n)))}{\textrm{max}(-(L_2(\gamma_n)))-\textrm{min}(-(L_2(\gamma_n)))}, 
	\end{split}	
\end{equation}
where $\alpha_l$ and $\alpha_h$ are the normalization weights of the positive and negative examples, respectively. $L_2(\cdot)$ means the $L_2$-norm distance. 

Last, we can obtain the dimension unimportance matrix $\beta$ as follows

\begin{equation}
	\beta\, =\,\sum_{1}^{\textrm{len}(\gamma_p)}\gamma_p[l]\cdot S_p[l] +\sum_{1}^{\textrm{len}(\gamma_n)}\gamma_n[h]\cdot S_n[h].
\end{equation}

The detailed process of calculating the dimension unimportance matrix is shown in Algorithm 1.

\begin{algorithm2e}[tp]
	\caption{Optimize the perturbation dimension}
	\label{alg1}
	\begin{algorithmic}[1]
		\REQUIRE Image $x_0$ and ground truth $y_0$, hard-label model $f_K(\cdot)$, initial perturbation direction $\theta_0$ and corresponding $\lambda_0$
		\ENSURE Adversarial perturbation $\delta$
		\STATE Search threshold $t$ $\leftarrow$ binary search algorithm 
		\STATE Set the value greater than $t$ in $\beta $ to 0 and the less ones to 1$\leftarrow \textrm{Bin}(\beta,t)$ 
		\STATE Get dimension optimized perturbation  $\delta \leftarrow \lambda_0 \theta_0\textrm{Bin}(\beta,t)$
		\STATE \Return Adversarial perturbation $\delta$		
	\end{algorithmic}
\end{algorithm2e}

\subsection{Optimize Perturbation Dimension}
By generating the initial perturbation and calculating the dimension unimportance matrix, we obtain the initial perturbation $\theta_0\lambda_0$ and the dimension unimportance matrix $\beta$. $\beta$ reflects the unimportance of each dimension of the initial perturbation direction $\theta_0$, and the larger the value, the lower the importance of the dimension. In this step, we search for a threshold $t$ which satisfies the following conditions
\begin{equation}
	\left\{
	\begin{array}{cl}
		f(x_0+ \lambda_0 \theta_0 \textrm{Bin}(\beta,\xi))\,\,!=y_0),  &  \textrm{when}\,\,\xi>=t \, , \\
		f(x_0+ \lambda_0 \theta_0 \textrm{Bin}(\beta,\xi))\,\,=y_0),  &  \textrm{when}\,\,\xi<t \, , \\
	\end{array} \right.
\end{equation}
where $\textrm{Bin}(\beta,\xi)$ denotes an operation {setting the values in $\beta$ which are greater than $\xi$ to 0 and the values which are less than $\xi$ to 1. We use a binary search algorithm to find the threshold. First, we take the initial upper bound (denoted as $high$) and lower bound (denoted as $low$) of the binary search algorithm as $\textrm{min}(\beta)$ and $\textrm{max}(\beta)$ respectively. Then we can obtain $\xi = \frac{\textrm{min}(\beta)+\textrm{max}(\beta)}{2}$. Second, by judging whether $f(x_0+ \lambda_0 \theta_0 \textrm{Bin}(\beta,\xi))\,\,!=y_0)$ is still satisfied, we can update the values of $high$ and $low$. That is, $ high = mid$ if it is true, and $low = mid$ otherwise. Third, we iterate the above search process until $high-low>10^{-6}$, and then output $t=high$. 

After the threshold $t$ is obtained, we can get the adversarial perturbation $\delta$ by setting the dimensions that are unimportant to the initial perturbation to zero. That means $\delta$ is $L_0$-norm optimization and can be expressed as
\begin{equation}
	\delta=\lambda_0 \theta_0 \textrm{Bin}(\beta,t). \\
\end{equation}
The detailed process of optimizing the perturbation dimension and generating the final adversarial example is shown in Algorithm 2. Through the above process, our method can generate adversarial examples $x=x_0+\delta$ with low $L_2$-norm distortion and fewer perturbed pixels. The overall procedure of our proposed method is shown in Algorithm 3.

\begin{algorithm2e}[tp]
	\caption{$L_p$-norm distortion-efficient adversarial attack}
	\label{alg1}
	\begin{algorithmic}[1]
		\REQUIRE Image $x_0$ and ground truth $y_0$, hard-label model $f_K(\cdot)$, baseline query number $N$, max query number $Q$
		\ENSURE  Adversarial example $x$
		\FOR{$b=1,2,...,N$ } 
			\STATE Generate initial direction vector $\theta_0$ and distance $\lambda_0$ using the same search algorithm in \cite{ChengSCC0H20}
		\ENDFOR 	 		
		\FOR{$r=1,2,...,Q-N$ } 
			\STATE Randomly generate 0/1 matrix
			\STATE Compute dimension unimportance matrix 
			$\beta$ $\leftarrow$ Algorithm 1				
		\ENDFOR 
		\STATE Get dimension optimized perturbation $\delta \leftarrow$ Algorithm 2
		\STATE \Return Adversarial example $x = x_0 + \delta$		
	\end{algorithmic}
\end{algorithm2e}

\section{Experimental Results}
\subsection{Datasets and Target Models}
We evaluate our algorithm on three different standard datasets, which are MNIST \cite{lecun1998gradient}, CIFAR10 \cite{krizhevsky2010cifar}, and ImageNet \cite{deng2009imagenet}. For a fair comparison, we use the CNN networks provided by \cite{carlini2017towards} and ResNet-50 \cite{he2016deep} provided by Torchvision \cite{marcel2010torchvision}. The details about implementation, training, and parameters refer to \cite{carlini2017towards}. Four leading hard-label black-box attack methods are adopted for comparison, which are Boundary attack \cite{brendel2018decision}, Opt-based attack \cite{cheng2019query}, Guessing Smart Attack \cite{brunner2019guessing}, Sign-OPT attack \cite{ChengSCC0H20}, and Bayes attack \cite{shukla2021simple}. 

\subsection{Metrics and Parameters Selection}
For each attack, we randomly select 100 examples from the validation set and generate adversarial perturbations for them. For untargeted attacks, we only consider examples that are correctly predicted by the target model. Three metrics are adopted to evaluate the black-box adversarial attacks: 1) Success Rate (SR). 2) $L_2$-norm loss (medium distortion). 3) $L_0$-norm loss (i.e. Percentage of Unattacked Pixels). SR is a percentage of the number of examples that have achieved an adversarial perturbation below a given threshold $\epsilon$ with less than $x$ queries. 
Our attack starts from generating a random Gaussian noise, and we can always obtain an initial adversarial example among hundreds of samples. In our experiments of $L_2$-norm setting, the thresholds take 1.5 for MNIST, 0.5 for CIFAR10, and 3.0 for ImageNet, respectively. Note that all images are normalized to [0,1] before they are passed to the model, i.e. the maximum pixel value is 1.
Because the image sizes of different datasets are different, we use the proportion of attacked pixels as the $L_0$-norm loss. Clearly, the fewer the median $L_2$ loss, the better performance. Similarly, a smaller $L_0$ loss indicates the adversarial example modifies fewer pixels, and the higher the attack success rate, the better performance of the attack method. Note that we choose the same set of examples for all the attacks for a fair comparison. 

\begin{figure}[tp]
	\subfloat[]{
		\begin{minipage}{0.33\linewidth}   
			\centering	
			\includegraphics[width=1\textwidth]{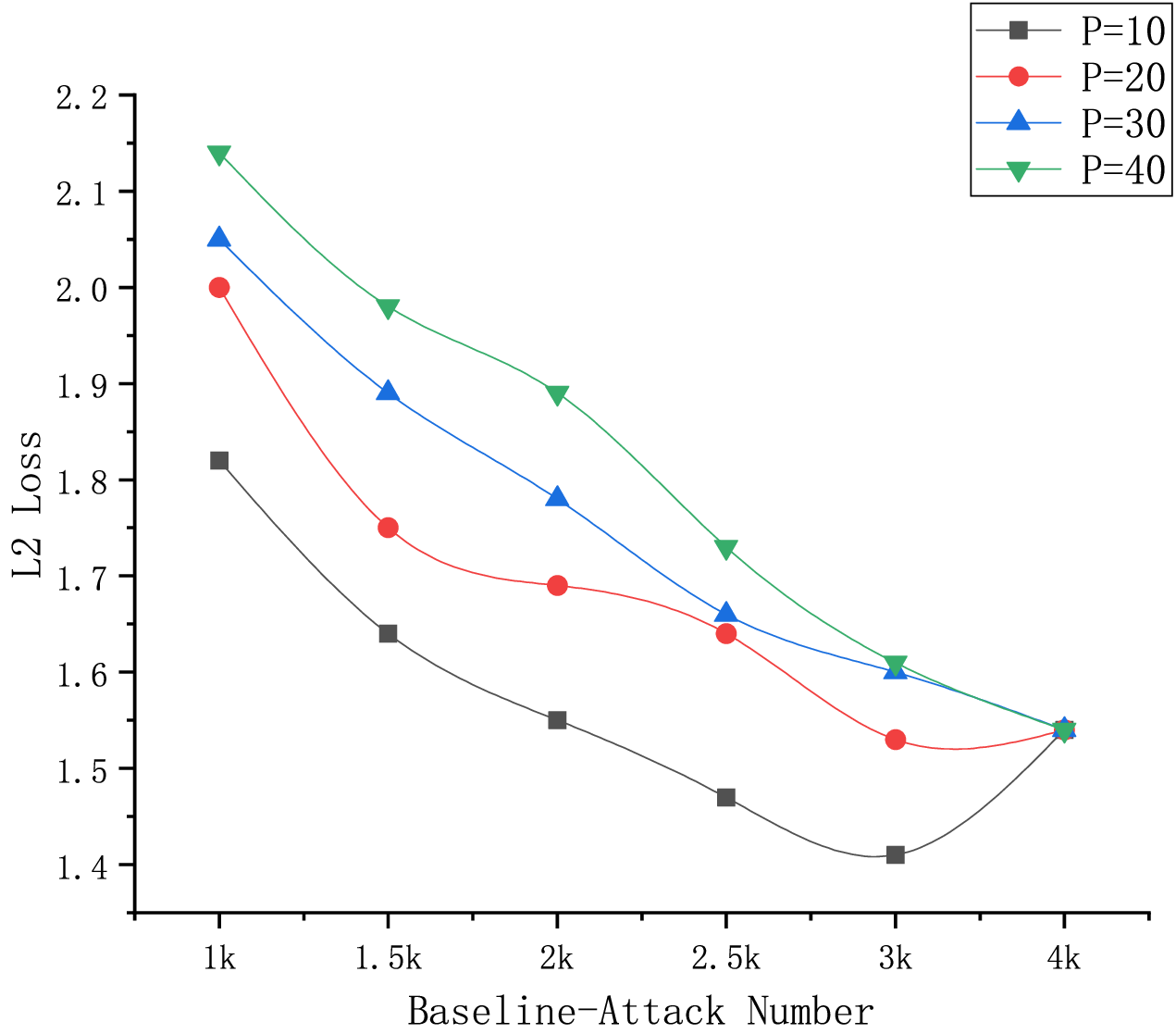}
		\end{minipage}
	}
	\subfloat[]{
		\begin{minipage}{0.33\linewidth}
			\centering
			\includegraphics[width=1\textwidth]{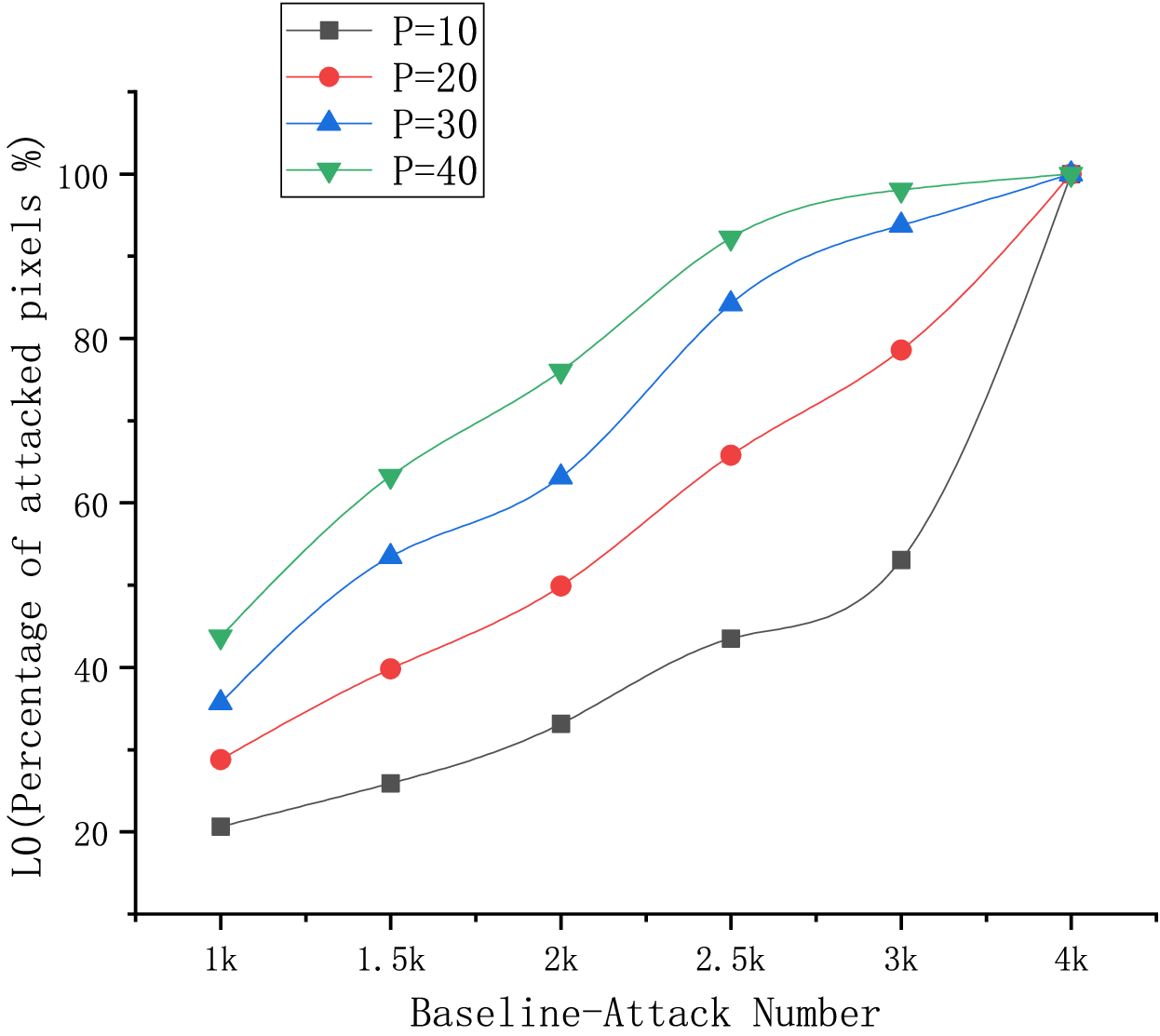}
		\end{minipage}
	}
	\subfloat[]{
	\begin{minipage}{0.33\linewidth}
		\centering
		\includegraphics[width=1\textwidth]{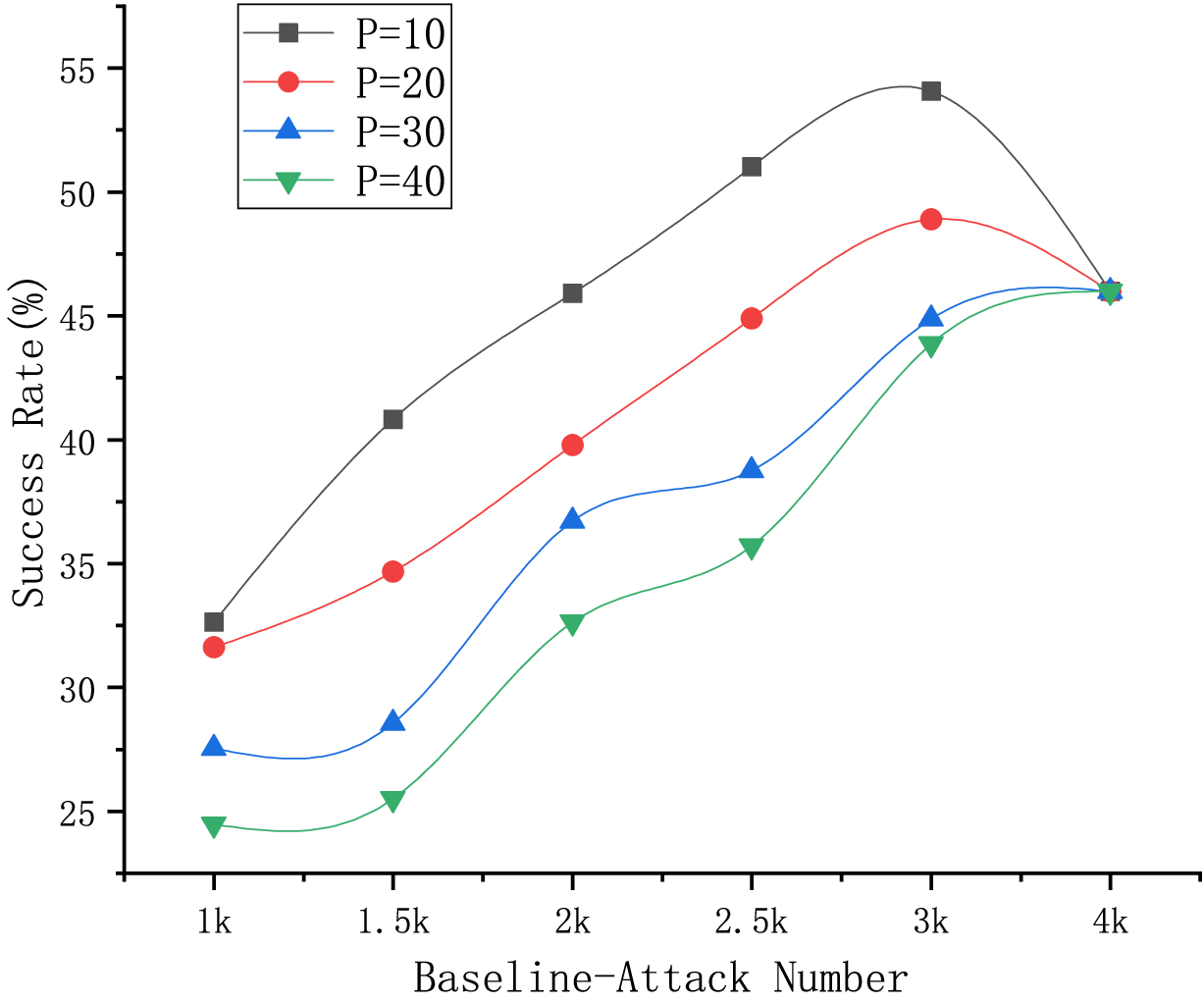}
	\end{minipage}
}
\quad
	\subfloat[]{
		\begin{minipage}{0.33\linewidth}   
			\centering	
			\includegraphics[width=1\textwidth]{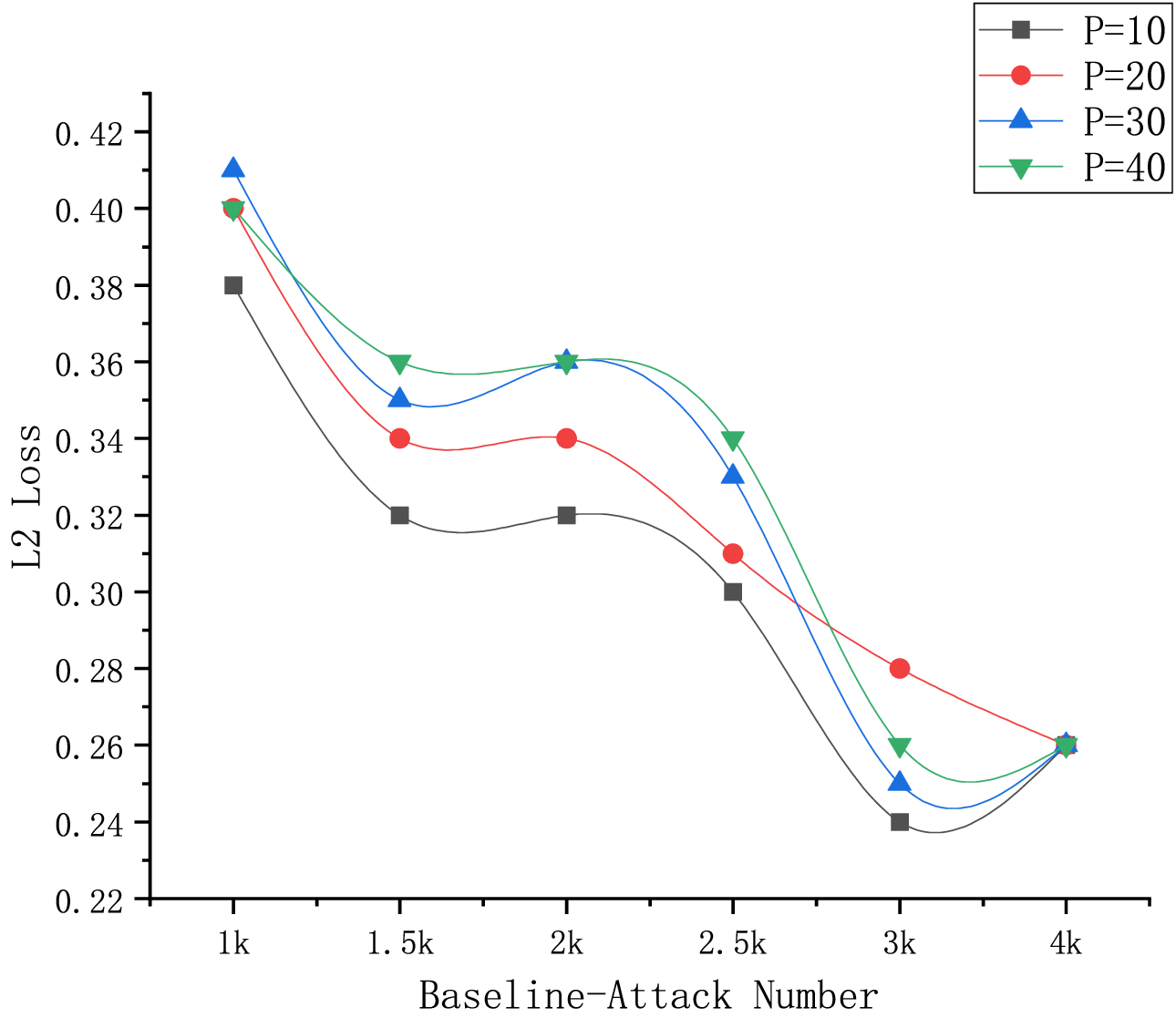}
		\end{minipage}
	}
	\subfloat[]{
		\begin{minipage}{0.33\linewidth}
			\centering
			\includegraphics[width=1\textwidth]{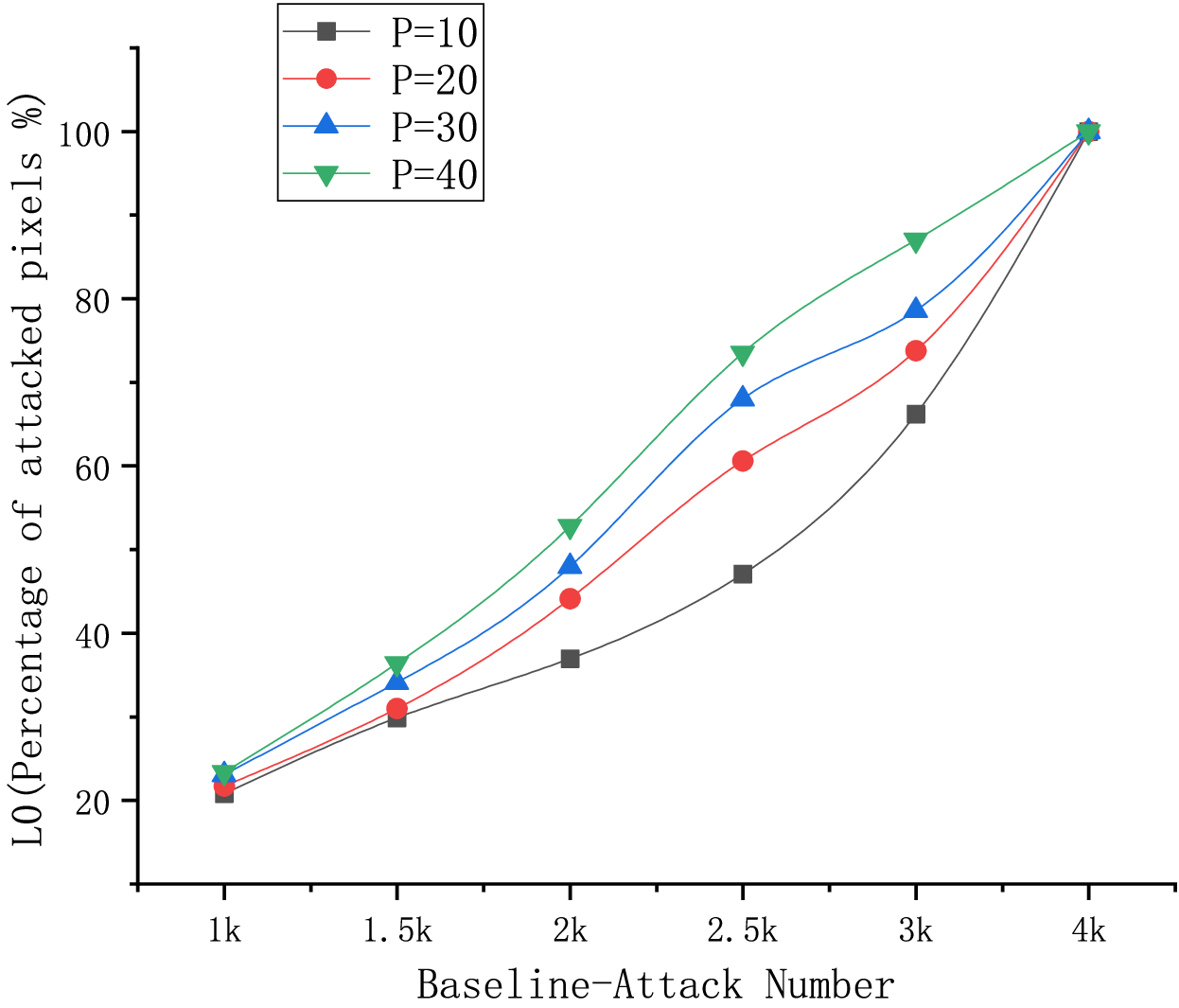}
		\end{minipage}
	}
	\subfloat[]{
	\begin{minipage}{0.33\linewidth}
		\centering
		\includegraphics[width=1\textwidth]{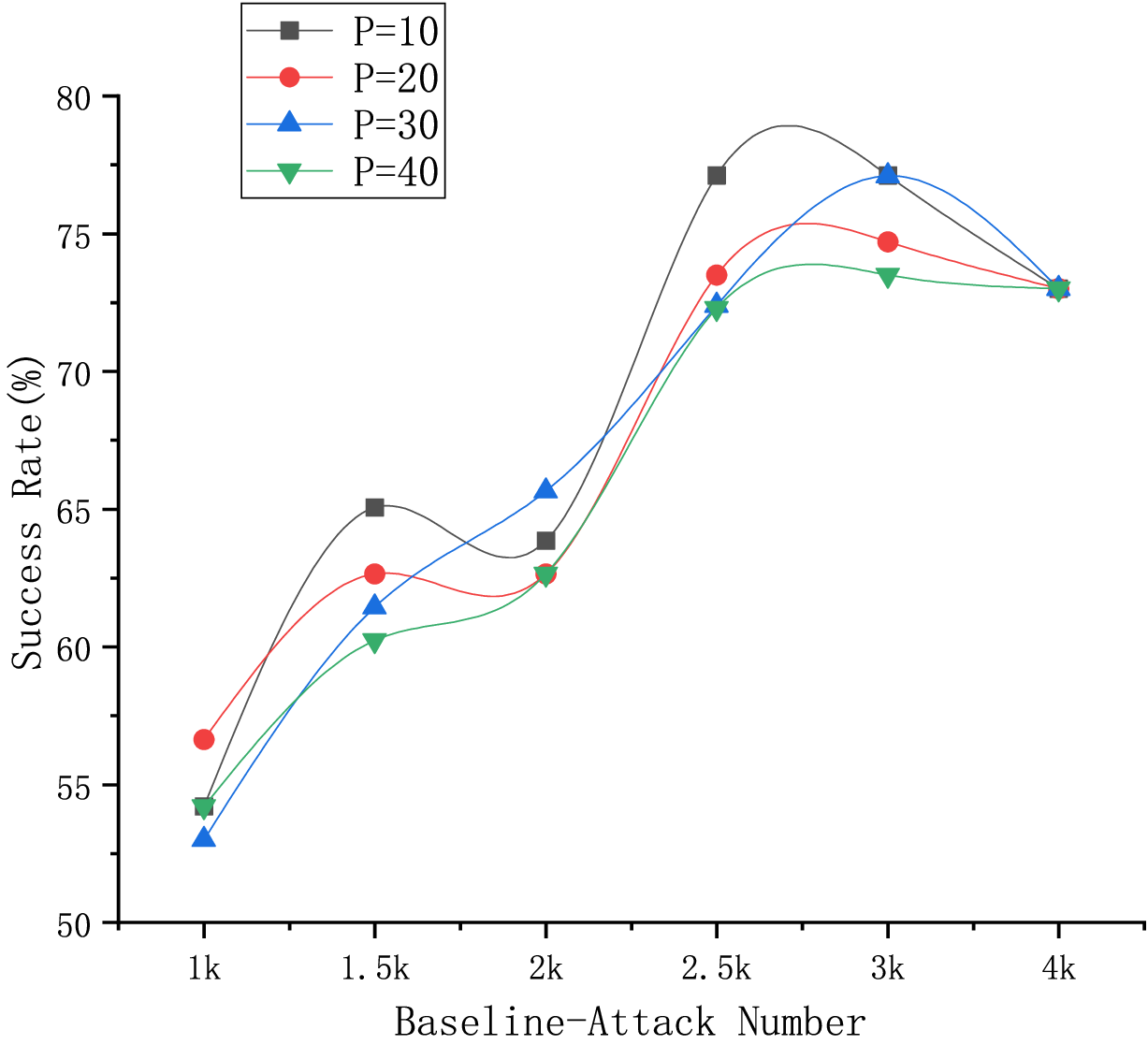}
	\end{minipage}
}
\quad
\subfloat[]{
	\begin{minipage}{0.33\linewidth}   
		\centering	
		\includegraphics[width=1\textwidth]{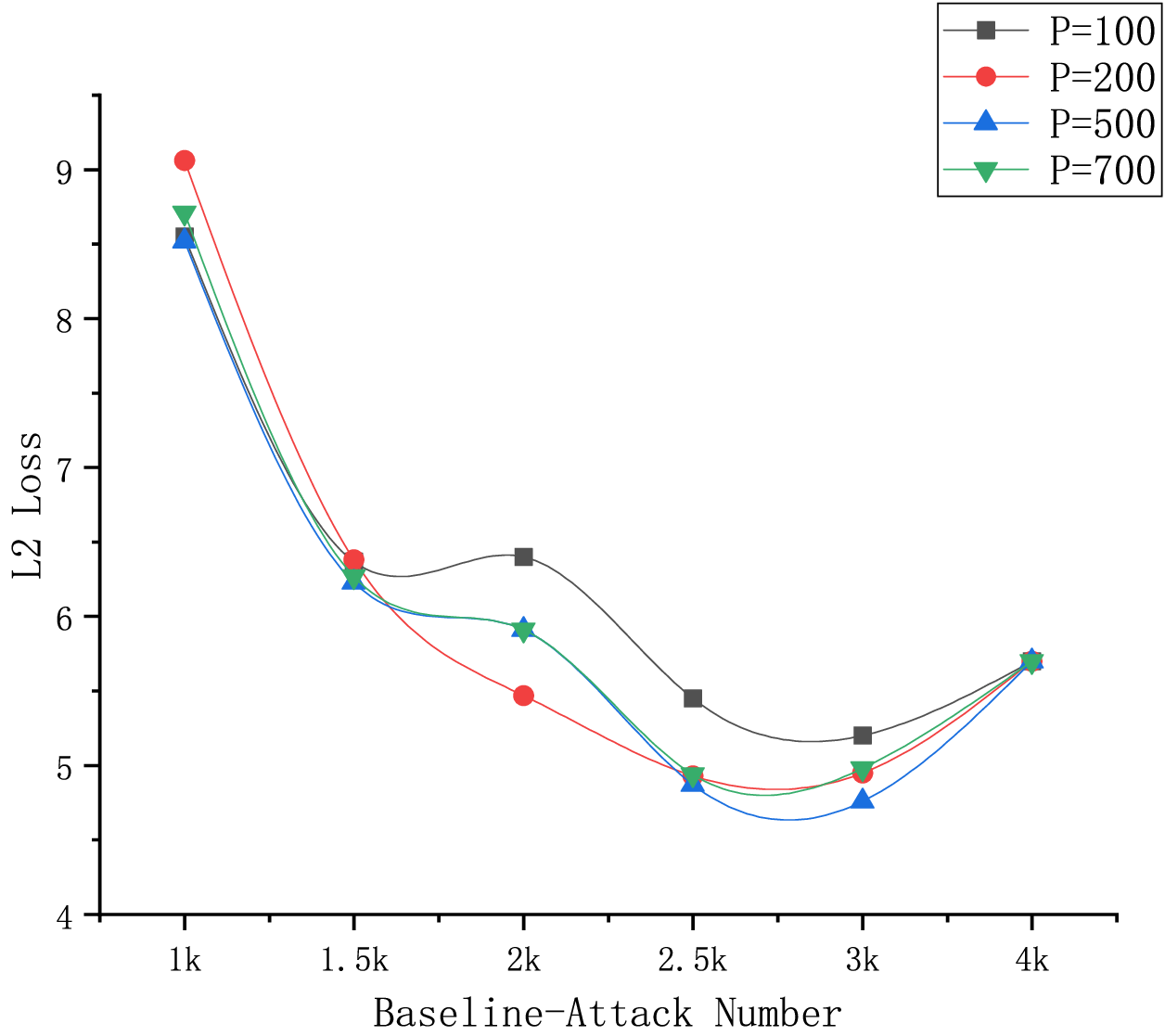}
	\end{minipage}
}
\subfloat[]{
	\begin{minipage}{0.33\linewidth}
		\centering
		\includegraphics[width=1\textwidth]{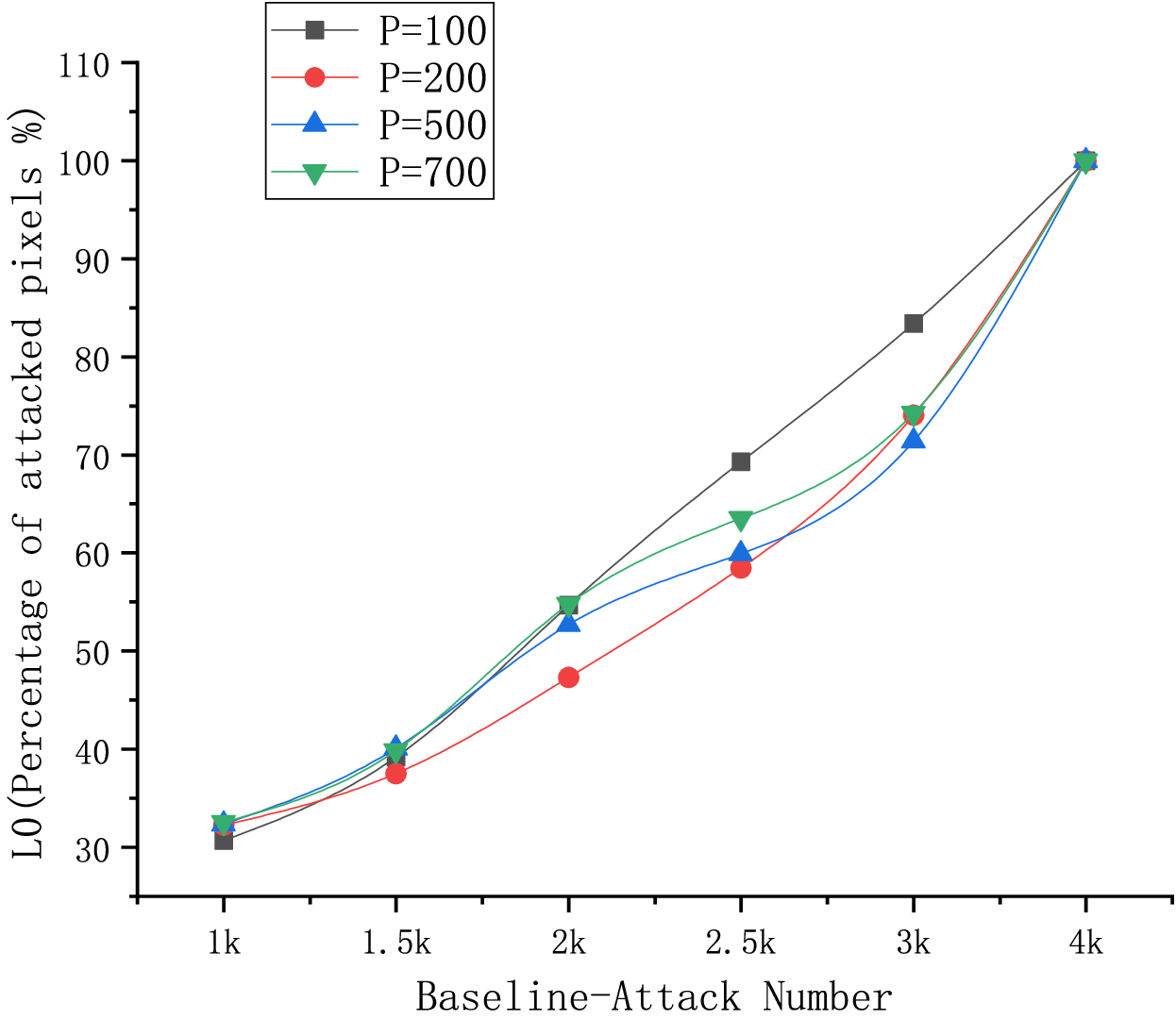}
	\end{minipage}
}
\subfloat[]{
	\begin{minipage}{0.33\linewidth}
		\centering
		\includegraphics[width=1\textwidth]{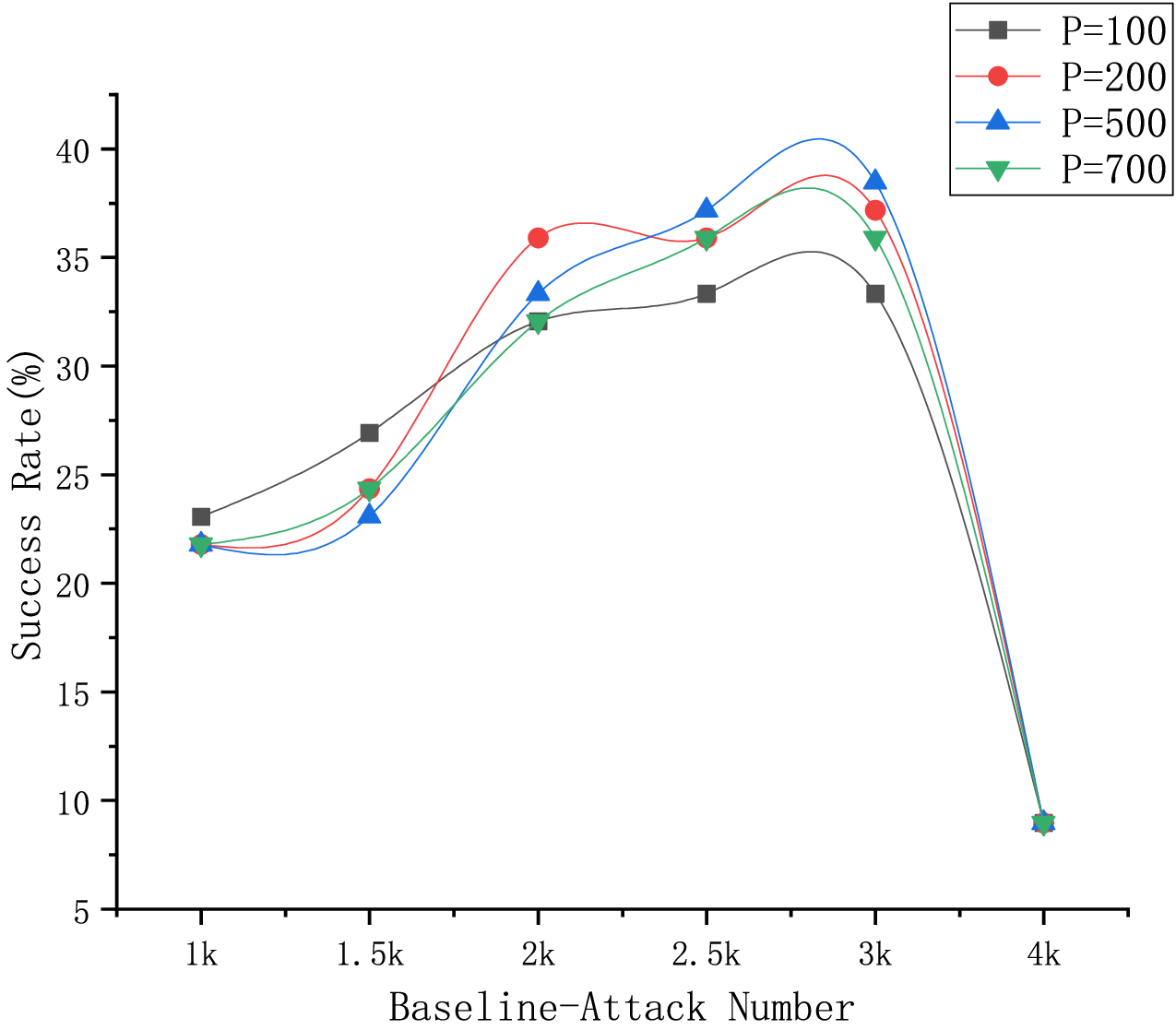}
	\end{minipage}
}
	\caption{Attack statistic curves using different super parameters $P$ and Baseline-Attack Number. The images in the first, second and third rows are the test results under MNIST, CIFAR10 and ImageNet, respectively.}
	\label{scatter}
	\vspace{-0.5cm}	
\end{figure}

According to the experiments, we find that the query number of the baseline and the number of dimensions (denoted as $P$) that are randomly set to zero in each iteration when calculating the dimension unimportance matrix can directly affect the attack effect. For each dataset, we test four values for $P$ respectively, which are 10, 20, 30, and 40 for MNIST and CIFAR10, and 100, 200, 500, and 700 for ImageNet. For each $P$ value, we take 6 values of the baseline query number within $[0, 4000]$, then count the attack performance, i.e. $L_2$ loss, $L_0$ loss, and success rate. For all the attacks, the maximum limited number of queries is set to 4000. 
Our method uses the Sign-OPT attack as the baseline, and we first use the Sign-OPT attack to generate an initial perturbation direction and the corresponding distance is shown in Step 2 of Algorithm 3. Then, we optimize the perturbation dimension. Therefore, when the query number of the baseline is the same as the maximum limited number, the attack is actually the Sign-OPT attack.
The results of all the combinations are shown in Figure 3. There are four curves corresponding to four different values of $P$. It can be seen from Figure 3 that with the increase of the baseline query number, the $L_2$ loss decreases first and then increases. For MNIST and CIFAR10, when the baseline query number is 3,000 and $P=10$, the $L_2$ loss gets the minimum value, and for ImageNet, the least $L_2$ loss occurs at the baseline query number is 3,000 and $P=500$. Unlike $L_2$ loss, the $L_0$ loss gradually increases with the baseline increase until 100$\%$ when it turns to the Sign-OPT attack. When the baseline query number is the same, the value of $P$ with the best loss of $L_0$ is respectively 10 on MNIST and CIFAR10 datasets and 500 on ImageNet datasets. Besides, it can be seen from the SR curves that, under the maximum limit of 4,000 total queries, the success rate can be improved by appropriately increasing the number of baseline queries. Similarly, SR gets the best value when $P=10$ for MNIST and CIFAR10, $P=500$ for ImageNet.

According to the experiment, we find that the $L_2$ loss and the $L_0$ loss show an opposite trend with the baseline change. To achieve the minimum $L_2$ loss, it is most appropriate to set the value of baseline query number to 3000 and $P=10$ for MNIST and CIFAR10, $P=500$ for ImageNet when the maximum limit of total queries is 4,000. If the value of the baseline query number is too large or too small, the $L_2$ loss will become worse. In this setting, we can also optimize the $L_0$ loss based on minimizing L2 loss. The percentages of attacked pixels on the three datasets are 53.2$\%$ for MNIST, 66.21$\%$ for CIFAR10, and 71.41$\%$ for ImageNet, respectively. Therefore, we take the value of the baseline query number as three-quarters of the maximum limit of total queries. This is because the baseline mainly optimizes the $L_2$ loss, which can change the perturbation of each pixel, but it cannot reduce the number of attacked pixels. In contrast, dimension optimization can be used to reduce the number of attacked pixels but not change the perturbation of the attacked pixels. Therefore, properly allocating the total number of queries can give full play to both advantages, achieving an imperceivable and sparse attack effect.

\begin{table*}[tp]
	\caption{Statistical comparison of various methods in untargeted attack setting.}
	\vspace{-0.2cm}
	\label{tab:freq}
	\resizebox{\textwidth}{!} 
	{
		\setlength\tabcolsep{2pt}
		\begin{threeparttable}
			\begin{tabular}{cccc|ccc|ccc}
				\hline
				\multirow{2}{*}{Attack} & \multicolumn{3}{c|}{MNIST} & \multicolumn{3}{c}{CIFAR10} &\multicolumn{3}{|c}{ImageNet (ResNet-50)} \\

				& Queries & M-$L_2$ &SR ($\epsilon=1.5$) & Queries & M-$L_2$ &SR ($\epsilon=0.5$)& Queries & M-$L_2$ &SR ($\epsilon=3.0$)\\
				\hline						
				\multirow{3}{*}{BA\cite{brendel2018decision}} & 4000&4.24 &1.0$\%$ &4000&3.12& 2.3$\%$ &4000 &209.63&0$\%$ \\
									     & 8000&4.24 &1.0$\%$ &8000&2.84& 7.6$\%$ &30000&17.40&16.6$\%$\\				
										& 14000&2.13 &16.3$\%$ &12000&0.78& 29.2$\%$ &160000&4.62&41.6$\%$\\				
				\hline

				\multirow{3}{*}{OPT\cite{cheng2019query}} & 4000&3.65 &3.0$\%$ &4000&0.77& 37.0$\%$ &4000 &83.85&2.0$\%$ \\
											& 8000&2.41 &18.0$\%$ &8000&0.43& 53.0$\%$ &30000&16.77&14.0$\%$\\				
										& 14000&1.76 &36.0$\%$ &12000&0.33& 61.0$\%$ &160000&4.27&34.0$\%$\\				
				\hline

				\multirow{3}{*}{GS\cite{brunner2019guessing}} & 4000&1.74 &41.0$\%$ &4000&0.29& 75.0$\%$ &4000 &16.69&12.0$\%$ \\
											& 8000&1.69 &42.0$\%$ &8000&0.25& 80.0$\%$ &30000&13.27&12.0$\%$\\				
										& 14000&1.68 &43.0$\%$ &12000&0.24& 80.0$\%$ &160000&12.88&12.0$\%$\\				
				\hline

				\multirow{3}{*}{Sign\cite{ChengSCC0H20}} & 4000&1.54 &46.0$\%$ &4000&0.26& 73.0$\%$ &4000 &5.71&8.0$\%$ \\
										& 8000&1.18 &84.0$\%$ &8000&0.16& 90.0$\%$ &30000&2.99&50.0$\%$\\				
										& 14000&1.09 &94.0$\%$ &12000&0.13& 95.0$\%$ &160000&1.21&90.0$\%$\\				
				\hline

				\multirow{3}{*}{Ours} & 4000&\textbf{1.41} &\textbf{54.1$\%$} &4000&\textbf{0.24}& \textbf{77.11$\%$} &4000 &\textbf{4.76}&\textbf{38.46$\%$} \\
					& 8000&\textbf{1.16} &82.7$\%$ &8000&\textbf{0.16}& \textbf{92.8$\%$} &30000& \textbf{1.25}& \textbf{91.0$\%$ }   \\				
					& 14000&\textbf{1.08} &90.8$\%$ &12000& 0.14 & \textbf{95.2$\%$} &160000&\textbf{0.32}  &\textbf{100$\%$ }\\				
				\hline
				
			\end{tabular}
			\begin{tablenotes}
				\footnotesize
				\item Abbreviated by BA: Boundary attack, GS: Guessing Smart, Sign: Sign-OPT, M-$L_2$: Median $L_2$ loss.
			\end{tablenotes}
		\end{threeparttable}
		
	}
\end{table*}

\subsection{Experimental Results on Untarget Attack }
Table 1 shows the experimental results on three datasets for the untarget attack. We mainly compare median $L_2$ loss (M-$L_2$) and SR when the maximum limit of the total query number is the same. From Table 1, we can see that compared with the four hard-label black-box attack methods, our method achieves a better attack effect. Especially for the ImageNet dataset, our method can reduce median $L_2$ loss and improve the attack success rate under different total attack numbers. In terms of MNIST, the median $L_2$ losses of our method are improved based on the baseline Sign-OPT. Although the success rate of our method remains the same level as the Sign-OPT method, it greatly improves over Sign-OPT (namely $8.1\%$) when the total attack number is set to 4000. As for CIFAR10, our method's success rate is higher than the other four methods in different total attack numbers. What's more, the median $L_2$ losses of our method are either better or competitive compared to Sign-OPT but obviously superior to the other three methods.  

\begin{table*}[tp]
	\caption{Experimental results for total untargeted attack.}
	\vspace{-0.2cm}
	\label{tab:freq}
	\resizebox{\textwidth}{!}
	{
		\setlength\tabcolsep{2pt}
		\begin{threeparttable}
			\begin{tabular}{cccc|cccc|cccc}
				\hline
				\multicolumn{4}{c|}{MNIST} & \multicolumn{4}{c}{CIFAR10} &\multicolumn{4}{|c}{ImageNet (ResNet-50)} \\
				
				Queries & M-$L_2$ &SR ($\epsilon=1.5$)& PP & Queries & M-$L_2$ &SR ($\epsilon=0.5$)& PP& Queries & M-$L_2$ &SR ($\epsilon=3.0$)& PP\\
				\hline						
				
				4000&1.41 &54.1$\%$& \textbf{46.9$\%$} &4000&0.24& 77.11$\%$&\textbf{ 33.79$\%$ }&4000 &4.76&38.46$\%$& \textbf{28.59$\%$ }\\
				8000&1.16 &82.7$\%$ & \textbf{39.0$\%$} &8000&0.16& 92.8$\%$& \textbf{18.2$\%$ }&30000& 1.25&91.0$\%$ & \textbf{14.0$\%$ }  \\				
				14000&1.08 &90.8$\%$&\textbf{30.3$\%$} &12000& 0.14 & 95.2$\%$ &\textbf{7.5$\%$} &160000&0.32  &100$\%$& \textbf{10.2$\%$}\\				
				\hline
				
			\end{tabular}
			\begin{tablenotes}
				\footnotesize
				\item Abbreviated by BA: Boundary attack, GS: Guessing Smart, Sign: Sign-OPT, M-$L_2$: Median $L_2$ loss, PP: Percentage of unattacked pixels.
			\end{tablenotes}
		\end{threeparttable}
		
	}
\end{table*}

According to Table 1, our method's $L_2$ loss and success rate are much better than the other four methods, including the baseline Sign-OPT, which are all designed with $L_2$ loss as the optimization function. The objective function proposed by our method requires that the $L_0$ loss is the minimum to minimize the $L_2$ loss, i.e., adversarial examples have the minimum $L_2$ distance and the least attacked pixels. This indicates that our attack method is more threatening and imperceptible.

Table 2 shows the whole untarget attack experimental results of our method on the three datasets. Compared to Table 1, Table 2 adds the $L_0$ loss. To better illustrate the optimization of our method in reducing the number of attacked pixels, the $L_0$ loss is represented by the percentage of unattacked pixels (denoted as PP). Table 2 shows that our method greatly reduces the number of attacked pixels based on $L_2$ optimization. Especially for the MNIST dataset, the effect is most obvious. When the total attack numbers are 4000, 8000, and 14000, respectively, our method improves by $46.9\%$, $39.0\%$, and $30.3\%$, compared with the baseline Sign-OPT which attacks the whole image. As the total attack number increases, the number of attacked pixels reduced by our method decreases accordingly. That is because estimating image gradient by the baseline is more accurate as the number of attacks increases. Thus, the dimensions that can be optimized are reduced. When the number of model queries is limited, and the estimation of image gradient is not accurate enough, our method shows a greater advantage. In terms of CIFAR10 and ImageNet, the same characteristics are shown, which further validates our analysis.

\begin{figure}[tp]
\centering
\begin{minipage}{0.3\linewidth}
	\centering
	\includegraphics[width=1\linewidth]{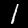}
\end{minipage}
\begin{minipage}{0.3\linewidth}
	\centering
	\includegraphics[width=1\linewidth]{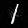}
\end{minipage}
\begin{minipage}{0.3\linewidth}
	\centering
	\includegraphics[width=1\linewidth]{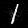}
\end{minipage}
\qquad
\begin{minipage}{0.3\linewidth}
	\centering
	\includegraphics[width=1\linewidth]{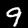}
\end{minipage}
\begin{minipage}{0.3\linewidth}
	\centering
	\includegraphics[width=1\linewidth]{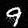}
\end{minipage}
\begin{minipage}{0.3\linewidth}
	\centering
	\includegraphics[width=1\linewidth]{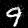}
\end{minipage}
\qquad
\begin{minipage}{0.3\linewidth}
	\centering
	\includegraphics[width=1\linewidth]{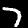}
\end{minipage}
\begin{minipage}{0.3\linewidth}
	\centering
	\includegraphics[width=1\linewidth]{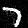}
\end{minipage}
\begin{minipage}{0.3\linewidth}
	\centering
	\includegraphics[width=1\linewidth]{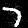}
\end{minipage}
\qquad
\begin{minipage}{0.3\linewidth}
	\centering
	\includegraphics[width=1\linewidth]{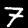}
\end{minipage}
\begin{minipage}{0.3\linewidth}
	\centering
	\includegraphics[width=1\linewidth]{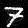}
\end{minipage}
\begin{minipage}{0.3\linewidth}
	\centering
	\includegraphics[width=1\linewidth]{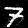}
\end{minipage}
\qquad	
\begin{minipage}{0.3\linewidth}
	\centering
	\includegraphics[width=1\linewidth]{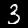}
	\caption*{Original images}
\end{minipage}
\begin{minipage}{0.3\linewidth}
	\centering
	\includegraphics[width=1\linewidth]{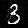}
	\caption*{Sign-OPT's}
\end{minipage}
\begin{minipage}{0.3\linewidth}
	\centering
	\includegraphics[width=1\linewidth]{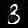}
	\caption*{Ours}
\end{minipage}

\vspace{-0.2cm}

\caption{Visual effect of different attack methods on MNIST. Images in the first column are the original images, the second column shows the adversarial examples of the Sign-OPT attack, and the third column shows the adversarial examples generated by our method on the same original image. Zoom in to see details.}
\vspace{-0.5cm}
\end{figure}

	\begin{table}[tp]
	\caption{Results for $L_2$ untargeted attacks on ImageNet classifiers with a query budget 1000.}
	\vspace{-0.2cm}
	\centering
	\label{tab:freq}
	\fontsize{9}{9}\selectfont
	\setlength{\tabcolsep}{3.9mm} 
	{
		\setlength\tabcolsep{2pt}
		\begin{threeparttable}
			\begin{tabular}{c|c|c|c|c|cc}
				\hline
				\multirow{2}{*}{Classifier} & \multirow{2}{*}{Method}  &\multirow{2}{*}{SR ($\epsilon=5.0$)}&\multirow{2}{*}{SR ($\epsilon=10.0$)}&
				\multirow{2}{*}{SR ($\epsilon=20.0$)}  &\multirow{2}{*}{$L_0$ (PP)}\\
				&&&&&&\\
				\hline										
				\multirow{3}{*}{ResNet50}&Sign-OPT~\cite{ChengSCC0H20}&13.74\%&24.68\%&43.51\%&-\\
				&Bayes attack~\cite{shukla2021simple}&\textbf{20.10\%}&\textbf{37.15\%}&66.67\%&-\\
				&Ours&16.83\%&32.25\%&\textbf{67.17\%}&\textbf{54.13\%}\\	
				\hline	
				\multirow{3}{*}{VGG16-bn}&Sign-OPT~\cite{ChengSCC0H20} &19.81\%&35.80\%&60.63\%&-\\
				&Bayes attack~\cite{shukla2021simple}&24.04\%&43.46\%&71.99\%&-\\
				&Ours&\textbf{27.11\%}&\textbf{45.97\%}&\textbf{78.94\%}&\textbf{48.35\%}\\
				\hline					
				
			\end{tabular}
			
		\end{threeparttable}
		\vspace{-0.5cm}
	}
\end{table}	

Shukla \textit{et al.} \cite{shukla2021simple} have proposed a Bayes attack based on Bayesian Optimization (BO). It is efficient when the query budget is small. 
We compare our method against the Bayes attack and the baseline Sign-OPT. One thousand images are used for untargeted attacks in the $L_2$-norm setting to calculate the Success Rate. We compare the Success Rate when the success threshold takes three different values, which are 5.0, 10.0, and 20.0, respectively. We only compare the Success Rate with the Bayes attack since it formulates the attack as an optimization problem with a constraint on perturbation distance. The results are shown in Table 3. From Table 3, 
we can see that our method achieves a higher Success Rate than the Bayes attack and Sign-OPT attack under the three success thresholds when the target model is VGG16-bn. At the same time, our perturbation involves a part of pixels, and the percentage of unattacked pixels (PP) is 48.35\% for the VGG16-bn.
Although the Success Rate is lower than the Bayes attack when the success thresholds are 5.0 and 10.0 for the ResNet50, our method only attacks 45.87\% of the whole pixels. Therefore, our method has a better attack performance.

It can be seen from Tables 1 and 2, that the CNN model used for classifying the MNIST dataset has a better robustness than that used for the CIFAR10 dataset since the adversarial examples generated by the former have higher Median $L_2$-norm loss, and lower Success Rate when the number of allowed model queries is the same. However, the adversarial examples for the MINST classification model can have less $L_0$-norm loss, as shown in the PP (the higher, the better) column of Table 2. This will be meaningful because the images in MNIST are grayscale, and most areas are pure black. Perturbing a larger number of pixels will make the attack more visible. 
The Success Rate on ImageNet is relatively lower than the other two datasets when the number of allowed model queries is also set to 4,000. This is because the search domain for adversarial examples is large. Interestingly, our method can improve the attack Success Rate by reducing the $L_0$-norm loss compared to the baseline. Furthermore, we can see from Table 3 that the ResNet-50 is more robust than the VGG16-bn.

Figure 4 shows the visual effect of the Sign-OPT attack and our method on the CIFAR10 dataset. The first row shows the original images, the images in the second row are the visual restoration of adversarial examples generated by Sign-OPT, and the third row shows the corresponding adversarial examples generated by our method on the same original image. On the one hand, it can be seen that the adversarial examples generated by our attack method can achieve the same, even better visual effect as Sign-OPT. As far as we know, Sign-OPT is a state-of-the-art hard-label black-box attack based on $L_2$-norm. According to the visual comparison, our method fully exploits $L_2$ attacks and generates tiny perturbations. On the other hand, by comparing columns 2 and 3, it is easy to find that the adversarial examples generated by our method have a much smaller number of changed pixels. This is because our method optimizes the $L_0$ loss simultaneously. Our method not only achieves a certain degree of optimization on $L_0$-norm, but also does not produce excessive single-pixel change.

\begin{table}[tp]
	\caption{Statistical comparison in target attack setting.}
	\vspace{-0.2cm}
	\centering
	\label{tab:freq}
	\fontsize{9}{9}\selectfont
	\setlength{\tabcolsep}{3.9mm}
	{
		\begin{threeparttable}
			\begin{tabular}{c|cc|cc|cc}
				\hline
				\multirow{2}{*}{Attack} & \multicolumn{2}{c|}{MNIST (Q=8000)} & \multicolumn{2}{c}{CIFAR10 (Q=8000)} &\multicolumn{2}{|c}{ImageNet (Q=10000)} \\
				
				& M-$L_2$ &SR ($\epsilon=1.5$)  & M-$L_2$ &SR ($\epsilon=0.5$) & M-$L_2$ &SR ($\epsilon=3.0$)\\
				\hline						
				BA\cite{brendel2018decision}&1.97 &4.72$\%$ &2.12& 1.96$\%$  &51.42 & 0.37$\%$ \\			
				\hline
				OPT\cite{cheng2019query} &2.59 &4.62$\%$ &1.24& 15.44$\%$ &40.11&2.24$\%$ \\
				\hline
				Sign\cite{ChengSCC0H20}&1.75 &37.38$\%$ &0.36& 73.03$\%$  &\textbf{19.14}&5.66$\%$ \\	
				\hline
				
				\multirow{2}{*}{Ours} &\textbf{1.66} &\textbf{38.38$\%$} &\textbf{0.28} & \textbf{76.54$\%$} &20.40&\textbf{5.66$\%$}\\		
				\cline{2-7}
				&\multicolumn{2}{c|}{PP = \textbf{29.34$\%$}} &\multicolumn{2}{c|}{PP = \textbf{8.53$\%$}} &\multicolumn{2}{c}{PP=\textbf{1.76$\%$}}	\\
				\hline
				
			\end{tabular}
			\begin{tablenotes}
				\footnotesize
				\item \scriptsize Abbreviated by BA: Boundary attack, Sign: Sign-OPT, M-$L_2$: Median $L_2$ loss, Q: Queries, PP: Percentage of unattacked pixels.
			\end{tablenotes}
		\end{threeparttable}
		\vspace{-0.5cm}
	}
\end{table}	
\subsection{Experimental Results on Target Attack}
Table 4 shows the experimental results on three datasets of the target attack. Like the untarget attack setting, we also compare M-$L_2$ and SR. It can be seen from Table 4 that our method has a better performance. Specifically, for the MNIST and CIFAR10 datasets, our method can not only reduce median $L_2$ loss but improve the SR when the limits of total attack numbers are set the same. In addition, to minimize $L_2$ loss, our method reduces the proportion of disturbed pixels on these two datasets by $29.34\%$ and $8.53\%$ respectively. Regarding ImageNet, the median $L_2$ losses and success rate of our method are similar to those of the Sign-OPT method. Interestingly, the number of pixels attacked by our method is greatly reduced. 

\subsection{Experimental Results on $L_\infty$-norm Setting}
Our method is also suitable for optimizing $L_\infty$-norm distance and $L_0$-norm distance simultaneously. Table 5 shows the experimental results in the $L_\infty$-norm setting. For untarget attack, we set the success thresholds to 0.3 for MNIST, 0.03 for CIFAR10 and ImageNet; for target attack, the thresholds are 0.3 for MNIST, 0.03 for CIFAR10, and 0.3 for ImageNet, respectively. From Table 5, it can be seen that our method achieves a higher Success Rate and a lower $L_\infty$-norm distortion in most cases. More importantly, our method greatly increases the PP, meaning a lower $L_0$-norm distortion. Therefore, our method can also achieve the goal of optimizing both $L_\infty$-norm and $L_0$-norm distortion simultaneously.

\begin{table}[tp]
	\caption{Performance comparison among various methods in the $L_\infty$-norm setting.}
	\vspace{-0.2cm}
	\label{tab1}
	\resizebox{\textwidth}{!} 
	{
		\setlength\tabcolsep{2pt}
		\begin{threeparttable}
			\begin{tabular}{c|c|c|ccccc|cccc|c}
				\hline
				\multirow{2}{*}{Dataset} & \multirow{2}{*}{Objective}  &\multirow{2}{*}{Queries} &\multicolumn{5}{c|}{SR (\%)} &\multicolumn{4}{c|}{Median $L_\infty$ Loss} &PP\\
				
				&&&$\epsilon$& BA&OPT&Sign&Ours&BA&OPT&Sign&Ours&Ours\\
				\hline				
				
				\multirow{2}{*}{MNIST}&Untargeted&15000&0.3&\textbf{77.38}&23.21&64.62& 69.39&\textbf{0.2253}&0.3472&0.2456& 0.2692&\textbf{77.84\%} \\
				&Targeted&20000&0.3&24.42&22.51&27.73&\textbf{27.75} &0.4003&0.4066&\textbf{0.3772}&0.3917&\textbf{69.78\%}\\
				\hline
				
				\multirow{2}{*}{CIFAR10}&Untargeted&4000&0.03&20.58&26.71&42.89&\textbf{43.37}&0.0539&0.0621&0.0357&\textbf{0.0350}&\textbf{59.57\%} \\
				&Targeted&8000&0.03&3.79&7.10&21.81&\textbf{23.68} &0.0807&0.1142&\textbf{0.0533}&0.0587 &\textbf{78.27\%} \\
				\hline	
				
				\multirow{2}{*}{ImageNet}&Untargeted&4000&0.03&7.36&7.89&16.46&\textbf{20.51} &0.1418&0.2330&0.1265&\textbf{0.1178} & \textbf{68.80\%}\\
				&Targeted&10000&0.3&29.71&31.40&\textbf{44.48}& 40.19&0.3874&0.3700&0.3207&\textbf{0.3201}&\textbf{1.06\%}\\						
				\hline
			\end{tabular}
			\begin{tablenotes}
				\footnotesize
				\item \scriptsize Abbreviated by BA: Boundary attack, Sign: Sign-OPT, PP: Percentage of unattacked pixels.
			\end{tablenotes}
		\end{threeparttable}
		\vspace{-0.5cm}
	}
\end{table}
\section{Conclusion}
In this paper, we have developed an $L_p$-norm distortion-efficient adversarial attack algorithm for the hard-label black-box classifiers, which can simultaneously optimize the $L_2$ (or $L_\infty$) and $L_0$ losses of adversarial examples. Specifically, we first generate an initial perturbation that is optimized on $L_2$-norm (or $L_\infty$-norm). Then we design a new algorithm to calculate a dimension unimportance matrix for the initial perturbation, which indicates the unimportance of each dimension in the initial perturbation. On this basis, we find an adversarial threshold by using a binary search algorithm with few model queries. The experimental results show that our algorithm is better than or at least comparable to the state-of-the-art approaches. This research provides a new challenge to the robustness of the hard-label black-box classifiers.





\bibliography{1}{}
%
%
%


\end{document}